\documentclass[conference]{IEEEtran}

\IEEEoverridecommandlockouts
\usepackage{cite}
\usepackage{amsmath,amssymb,amsfonts}
\usepackage{multirow}
\usepackage{graphicx}
\usepackage{subfig}
\usepackage{textcomp}
\usepackage{xcolor}
\usepackage[utf8]{inputenc}
\usepackage[english]{babel}
\usepackage{fancyhdr}
\usepackage[font=footnotesize,skip=0pt]{caption}
\usepackage{array}
\usepackage{tabularx}
\usepackage[ruled,vlined]{algorithm2e}
\usepackage{algpseudocode}
\usepackage[section]{placeins}
\usepackage{float}

\usepackage{amsthm}

\def\BibTeX{{\rm B\kern-.05em{\sc i\kern-.025em b}\kern-.08em
    T\kern-.1667em\lower.7ex\hbox{E}\kern-.125emX}}
\begin{document}

\title{ Deep Unsupervised Learning for Generalized Assignment Problems: A Case-Study of User-Association in Wireless Networks      \\
}
\author{Arjun Kaushik,  Mehrazin Alizadeh, Omer Waqar, {\em  Member IEEE}, and~Hina~Tabassum, {\em Senior Member IEEE}
\thanks{ A. Kaushik, M. Alizadeh, and H.~Tabassum are with the Lassonde School of Engineering at York University, ON, Canada (e-mail: akarjun@my.yorku.ca, mehrazin@yorku.ca, hinat@yorku.ca).
O. Waqar is with the department of engineering at Thompson Rivers University, BC, Canada (e-mail: owaqar@tru.ca).}
\vspace{-8mm}
}

\raggedbottom

\maketitle

\begin{abstract}
There exists many resource allocation problems in the field of wireless communications which can be formulated as the generalized assignment problems (GAP). GAP is a generic form of linear sum assignment problem (LSAP) and is more challenging to solve owing to the presence of both equality and inequality constraints. We propose a novel deep unsupervised learning (DUL) approach to solve GAP in a time-efficient manner. More specifically, we propose a new approach that facilitates to train a deep neural network (DNN) using a customized loss function. This customized loss function constitutes the objective function and  penalty terms corresponding to both equality and inequality constraints. Furthermore, we propose to employ a $\mathsf{Softmax}$ activation function at the output of DNN along with tensor splitting which simplifies the customized loss function and guarantees to meet the equality constraint. As a case-study, we consider a typical user-association problem in a wireless network, formulate it as GAP, and consequently solve it using our proposed DUL approach. Numerical results demonstrate that the proposed DUL approach provides  near-optimal  results with significantly lower time-complexity.


\end{abstract}

\begin{IEEEkeywords}
Deep neural networks (DNNs), generalized assignment problem (GAP), unsupervised learning (UL), user-association and  wireless networks.
\end{IEEEkeywords}

\vspace{-3mm}
\section{Introduction}

A variety of resource allocation problems in wireless
communications can be modeled as generalized assignment
problems  (GAP), where the aim is to assign $n$ resources to $m$ agents in an optimum manner. Different from the linear sum assignment problems (LSAP) with equality constraints, GAP problems can handle  both equality and inequality constraints~\cite{abc}. GAP is a classical NP-hard combinatorial optimization problem and is widely applicable in wireless research problems, such as computation offloading in edge computing systems \cite{zarandi2021delay}, user scheduling with load balancing~\cite{8645483,tabassum2015spectral}, sub-channel assignment  \cite{7560605, 7417643}, antenna selection \cite{khalili2020joint}, etc.    For instance, in \cite{8645483},  a user association and load balancing problem was modeled as GAP where inequality constraints were applied to ensure that each user can be associated to only one base station (BS) at a time. Furthermore,  \cite{7560605} and \cite{7417643} applied GAP to a sub-channel assignment problem in which the inequality constraint is used to ensure that a sub-channel can only be assigned to a certain number of users at a time, alongside that each user can only occupy a certain number of sub-channels.  In \cite{7314981,7442793, 8538502}, the authors modeled the user association problem through GAP,  where the constraints are applied to ensure that the users served by a particular BS are served in BS clusters of the same size, and each user must be accepted by at most one BS, respectively. 

Most of the existing algorithms applied conventional non-data driven optimization methods to solve the aforementioned problems.  However, the computational complexity of such solutions is generally high which hinders the practicality of these solutions.  Furthermore, in emerging 5G/6G wireless networks,  the channel coherence time is much smaller for higher frequencies (e.g., mm-wave and THz), thus optimization needs to be performed quite often. This implies that a trivial exhaustive search method will be computationally prohibitive even for moderate size networks. In the sequel, artificial intelligent (AI)-enabled algorithms can potentially minimize the time complexity while enhancing the scalability.  

Recently, few research works have considered supervised and reinforcement  learning for solving GAP problems, such as user association problems \cite{9046032, 8645483,7724478, 7950979,91187,8405562}, channel assignment problems \cite{8967053, 8669028,1191240}, etc. 
Nonetheless, the performance of supervised learning (SL) rely on the quality of the labels that are generated via computationally-intensive algorithms. On the other hand, reinforcement learning (RL) algorithms are more suitable for problems that are formulated as Markov Decision Processes (MDPs) and their convergence for constrained optimization problems is not guaranteed.

In this paper, we propose a \textit{novel deep unsupervised learning (DUL)} approach to provide near optimal results for GAP problems.  Subsequently, there is no need to solve for the ground truth as there is no labeled data requirements.  Generally,  the  main  challenge  for any  DUL  approach  is to  implement  the  constraints  in the DNN architecture. For instance, transmit power constraint has been handled in  \cite{Lee2018a} and \cite{Liang2020} by using $\mathsf{Sigmoid}$ activation function at the output layer. However, as GAP involves intricate equality and inequality constraints, the architecture adopted in \cite{Lee2018a} and \cite{Liang2020} is not applicable. 

Different from the aforementioned works and in order to train the DNN through unsupervised learning, we made the following main contributions in this paper:  \textbf{(i)} we provide a new loss function which consists of an objective function and penalty terms corresponding to both equality and inequality constraints, \textbf{(ii)} we show that the loss function can be simplified by splitting an output layer into multiple tensors and each tensor is activated by a separate $\mathsf{Softmax}$ function. This simplified loss function makes sure that the equality constraint is always satisfied, and \textbf{(iii)} as a case-study, we consider a typical user-association problem in a wireless network, formulate it as GAP, and consequently solve it using our proposed DUL approach. We demonstrate that our proposed DUL approach has much lower time complexity as compared to the optimal solution obtained by CVX with a very high near optimal prediction accuracy.


The remainder of this paper is organized as follows. Mathematical representation of GAP is presented in Section II. Furthermore, the details of our proposed DUL approach are  provided in Section III. Section IV presents the case study of user association in a wireless network. Numerical examples and results of the DUL approach are also provided in this section. Finally, the paper is concluded in Section V.

\textit{Notations:} Scalars and vectors are denoted by italic and bold-face lower-case letters, respectively. $\mathsf{ReLU}(x)\triangleq x^{+}\triangleq \text{max}(x,0)$ represents a Rectified Linear Unit activation function. Moreover, $\mathbb{R}^{M\times1}$ denotes the space
of $M$-dimensional real-valued vector.

\section{Mathematical Representation of GAP}

A classical GAP deals with optimal assignment of $I$ items to $J$ knapsacks such that each item is assigned to only one knapsack without assigning to any knapsack a weight greater than its capacity \cite{Silvano1990}. Mathematically, GAP is formulated as an optimization problem ($\mathcal{P}$), given by

\begin{equation}
\begin{array}[b]{c}
\!\!\!\!\!\!\!\!\!\!\!\!\!\!\!\!\!\mathcal{P}:\,{\displaystyle \text{maximize}}\quad Z\triangleq\sum_{i=1}^{I}\sum_{j=1}^{J}u_{i,j}p_{i,j},\:\\
\\
\!\!\!\!\!\!\!\!\!\!\!\!\!\!\!\!\!\!\!\!\!\!\!\!\!\!\!\!\!\!\!\!\!\!\!\!\!\!\!\!\!\!\!\!\!\!\!\!\!\!\!\!\!\!\!\!\!\!\!\!\!\!\!\!\!\!\!\!\!\!\!\!\!\!\!\!\!\!\!\!\!\!\!\!\!\!\!\!\!\!\!\!\!\textrm{subject to,}\\
\\
\begin{array}[b]{c}
\!\!\!\!\!\!\!\!\!\!\!\!\!\!\!\!\!\mathrm{C1}:{\displaystyle \,\sum_{j=1}^{J}}u_{i,j}=1,\quad\forall i\in \mathcal{I}\triangleq\{1,2,..I\},
\end{array}\\
\mathrm{C2}:\,{\displaystyle \sum_{i=1}^{I}}w_{i,j}u_{i,j}\leq{\displaystyle c_{j}},\quad\forall j\in \mathcal{J}\triangleq\{1,2,..J\},\\
\\
\!\!\!\!\!\!\!\!\!\!\!\!\!\!\!\!\!\!\!\!\!\!\!\!\!\mathrm{\quad C3}:\,u_{i,j}\in\{0,1\},\quad\forall i\in \mathcal{I},\,\forall j\in \mathcal{J},
\end{array}
\label{eq: optimization_problem}
\end{equation}
where $c_{j}$ is the capacity of $j$th knapsack, $p_{i,j}$ and $w_{i,j}$ represent profit and weight of $i$th item, respectively when it is assigned to the $j$th knapsack. Moreover, $u_{i,j}=1$ if $i$th item is assigned to $j$th knapsack or equal to zero otherwise.   It is evident from eq. (\ref{eq: optimization_problem}) that the problem $\mathcal{P}$ becomes equivalent to the LSAP (which is analyzed in \cite{Lee2018}) only for a special case i.e., when $c_{j}=1$, $w_{i,j}=1$ $\forall i\in \mathcal{I},\, j\in \mathcal{J}$ and $I=J=n$. The LSAP is solved through an optimal Hungarian algorithm, which has a computational complexity of \textit{O}$\left(n^{3}\right)$ \cite{Silvano1990}.

\section{Proposed Deep Unsupervised Learning Approach}
Due to the fact that GAP is a generic form of LSAP, it is more challenging to solve this problem. Moreover, the DL approach adopted in \cite{Lee2018} cannot be directly applied. Considering this, contrary to \cite{Lee2018}, we propose a new DNN-based approach in which a single DNN learns the GAP directly without a need to generate time-consuming labels i.e.,  we train a DNN through unsupervised learning using customized loss function and tensor splitting. Further details of our approach are given as follows:

\subsection{Loss Function}
Leveraging the analytical expressions for the objective and constraints functions in eq. (\ref{eq: optimization_problem}) and for any given arbitrary values of $c_{j}$, $w_{i,j}$ $\forall i\in \mathcal{I},\, j\in \mathcal{J}$, we define a \textit{customized} loss function that is minimized through DNN. This customized cost function $L$ is given in (\ref{eq: original_cost}), shown at the top of next page,
\begin{figure*}[tbh]
\begin{equation}
    L = {\displaystyle \frac{1}{|\mathcal{F}|}}\sum_{\mathbf{f}\in\mathcal{F}}\left[-Z\left(\mathbf{f},\theta\right) + \lambda_{1} \cdot \left(\sum_{i = 1}^{I} \mathsf{ReLU}(1 - \sum_{j=1}^{J}u_{i,j})\right) + \lambda_{2} \cdot \left(\sum_{j = 1}^{J}\mathsf{ReLU}(c_{j} - \sum_{i=1}^{I}w_{i,j}u_{i,j})\right)\right].
\label{eq: original_cost}    
\end{equation}
\\
\begin{equation}
    L_{\text{simplified}} = {\displaystyle \frac{1}{|\mathcal{F}|}}\sum_{\mathbf{f}\in\mathcal{F}}\left[-Z\left(\mathbf{f},\theta\right)  + \lambda \cdot \left(\sum_{j = 1}^{J}\mathsf{ReLU}(c_{j} - \sum_{i=1}^{I}w_{i,j}u_{i,j})\right)\right]. 
\label{eq: simplified_cost}    
\end{equation}
\hrule
\end{figure*}
where $\mathbf{f}$ denotes a feature vector which is flattened to contain corresponding $p_{i,j}$ values. It is worth pointing out here that $u_{i,j}$  in (\ref{eq: original_cost}) are the values from the output of a DNN corresponding to $\theta$ which denotes the set of trainable network parameters for DNN. Moreover, $\mathcal{F}$  represents a mini-batch which contains certain number of examples for feature vectors, where the number of examples is determined by its size $|\mathcal{F}|$.  Furthermore, the terms $(\sum_{i = 1}^{I} \mathsf{ReLU}(1 - \sum_{j=1}^{J}u_{i,j}))$ and $(\sum_{j = 1}^{J}\mathsf{ReLU}(c_{j} - \sum_{i=1}^{I}w_{i,j}u_{i,j}))$ in  (\ref{eq: original_cost}) are incorporated in order to tackle the constraints C2 and C3, respectively. In other words, these two terms are considered as penalty terms providing an incentive to the DNN meeting the constraints. Furthermore, $\lambda_{1}$ and $\lambda_{2}$ are treated as hyper-parameters of the DNN, which implies that large values of these $\lambda$'s can cause a bias towards meeting the constraints (and neglecting the objective function) while very small values tend the DNN to ignore the constraints all together (and bias towards the objective function). The impact of these $\lambda$'s on the performance of DNN, which will be discussed in section IV.

\begin{figure*}
\centering
    \includegraphics[scale=0.4]{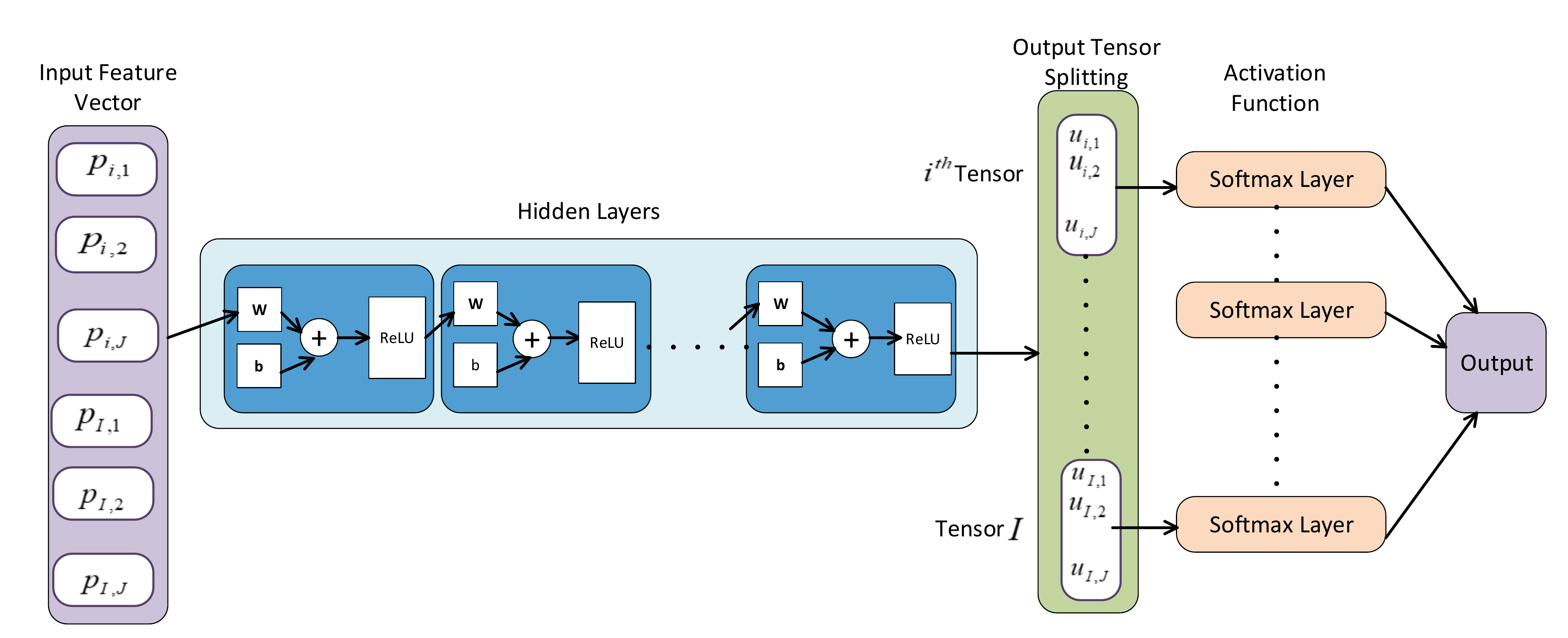}  
    \caption{Illustration of the proposed deep neural network architecture.}
  \label{fig:network_architecture}
\end{figure*}

\subsection{Network Architecture}
 A fully-connected (FC) feed-forward neural network is considered in which the hidden layers have $\mathsf{ReLU}$ activation functions, as shown in Fig. \ref{fig:network_architecture}. The DNN was trained using mini-batch gradient descent method in which each mini batch $\mathcal{F}$ is generated from independent feature vectors. The novelty in DNN's architecture lies in the final layer which is split into $I$ tensors, each having a $\mathsf{Softmax}$ activation function. There are two main advantages of this approach: (i) it guarantees to satisfy the constraint C2 and (ii) the loss function given in (\ref{eq: original_cost}) is simplified and can be rewritten in (\ref{eq: simplified_cost}), where for sake of brevity, we have taken $\lambda\triangleq\lambda_{1}$. \textit{To the best of our knowledge, this \textquoteleft tensor splitting' approach has not been used in any prior work.} The loss function of (\ref{eq: simplified_cost}) along with all the $u_{i,j}$'s (at the output of each $\mathsf{Softmax}$ layer) are then utilized during the training process of DNN.  

\subsection{Training Stage and Constraint Violation Probability}
 A feature vector, $\mathbf{f} \in$ $\mathbb{R}^{IJ\times1}$  is fed to the input of the proposed DNN. The $I$ tensors, each having $J$ values are used at the output layer to minimize the loss function given in (\ref{eq: simplified_cost}). For this minimization,\textquoteleft Adam' (adaptive moment estimation) \cite{kingma2014adam} optimization algorithm is used. It is worth mentioning here that $\lambda$ is one of the hyper-parameters and is chosen to obtain the balance between the constraint violation probability (i.e., the probability that measures a violation of constraint C3) and the maximization of objective function of problem $\mathcal{P}$, as will be explained in section IV. Note that,  first the constraint violation probability for each knapsack is calculated by dividing the number of examples (in the test data set) which do not meet constraint C3 for that particular knapsack by the total number of examples in the test  data set. After this, the constraint violation probability is averaged over all the knapsacks. As such, it is referred to as \textit{average constraint violation probability} in the paper.


\section{Case Study: User-Association in Wireless Networks}

In this section, we first describe a typical user-association problem in a wireless network and then show that this user-association problem can be formulated as GAP. Therefore, our proposed DUL based approach can readily be used to solve user-association problem in a time-efficient manner.

\subsection{System Setup}
A two tier downlink network consisting of RF BSs and THz BSs is considered. We assume that there exists a software-defined network (SDN) controller that performs user associations. Furthermore,  without loss of generality, the assumption of BSs and users being uniformly distributed in a circular region is made. The set of users is denoted by $\mathcal{I} = \{1,2,\cdots,I\}$  and $\mathcal{J} = \{1,2,...,J\}$ represents the set of BSs. Hence, the roles of items and knapsacks in section II are taken by users and BSs in this section.
\subsubsection{ RF Channel and SINR Model}  
The channel power of the $i$th user from the $j$th RF BS communication link is modeled as {$h_{R}= \gamma_{R} \rho_{i,j}^{-\alpha}\chi_{i,j}$}, where  $\gamma_{R} \triangleq {c^2}/{\left(4\pi f_{R}\right)^2} $, $f_{R}$ is the RF carrier frequency in GHz and $c$ is the speed of light i.e., $c = 3 \times 10^8$ m/s. Moreover, $\alpha$ is the path-loss exponent, $\chi_{i,j}$ and $\rho_{i,j}$  represent the exponentially distributed unit mean channel power and the distance between the $i$th user and $j$th BS, respectively. The RF BSs are equipped with omnidirectional antennas, therefore, for $i$th user which is served by $j$th RF BS, its corresponding  $\left(\textrm{SINR}^R_{i,j}\right)$ is given as:
\begin{equation}
\mathrm{SINR}^R_{i,j}= \frac{P_{R}h_{R}}{N_{0}+I_{\mathrm{agg}}^{R}},
\label{eq: SINR_RF} 
\end{equation}
where $P_{R}$ is the transmit power of all the RF BSs, $N_{0}$ is the power of the additive white Gaussian noise (AWGN) at the user. Furthermore, $\left(I_{\mathrm{agg}}^{R}\right)$ denotes the aggregate SINR at the $i$th user from the interfering RF BSs and is given as: 
\begin{equation}
I_{\mathrm{agg}}^{R} \triangleq {\displaystyle \sum_{\begin{array}{c}
\forall k\in I,\forall m\in J\\
\{k,m\}\neq\left\{ i,j\right\} 
\end{array}}}P_{R} \gamma_{R}\rho_{k,m}^{-\alpha}\chi_{k,m} 
\end{equation}


\subsubsection{THz Channel and SINR Model}  
Due to  high molecular absorption and the dense deployment, the  line-of-sight (LoS) transmissions are more dominant than that of non-line-of-sight (NLoS). Therefore, in this paper, we consider only the LoS transmission\footnote{{The consideration of NLoS with accurate reflection, scattering, and diffraction models deserves a separate study and has been left for future investigation.}} between users and THz BSs. The channel power of the $i$th user from the $j$th THz BS communication link is modeled as 
$h_{i,j} =\gamma_{T} \exp\left(-k_{a}r_{i,j}\right)/r_{i,j}^2$, where $\gamma_{T}\triangleq c^2/{\left(4\pi f_{T}\right)^2}$, $f_{T}$ is the operating frequency in THz, $k_{a}$ is the molecular absorption coefficient and $r_{i,j}$ is the distance between the $i$th user and $j$th BS. Moreover, the directional antennas gains are  modeled as \cite{di2015stochastic}, i.e.,
\begin{equation}
\label{eq:gain}
  G^{T}_{q} \left(\theta\right) =
    \begin{cases}
      G_{q}^{\left(\mathrm{max}\right)}, & \mid \theta_q \mid \leq w_{q}\\ 
      G_{q}^{\left(\mathrm{min}\right)}, & \mid \theta_q \mid > w_{q}
    \end{cases},  
\end{equation}
where $q\in \{\mathrm{tx,rx}\}$,
$G^{\mathrm{T}}_{\mathrm{tx}}\left(\theta\right)$ and $G^{\mathrm{T}}_{\mathrm{rx}}\left(\theta\right)$ represent the directional transmitter and receiver antenna gains, respectively. Furthermore, $\theta \in [-\pi,\pi)$ is the angle of the boresight direction, $w_{q}$ is the main lobe beamwidth, $G_{q}^{\left(\mathrm{max}\right)}$ and $ G_{q}^{\left(\mathrm{min}\right)}$ are beamforming gains of the main and side lobes, respectively. The typical user and its desired THz BS align such that their main lobes coincide through beam alignment techniques \cite{di2015stochastic}. With the assumption that the main lobe of $i$th user coincides with $j$th THz BS, the corresponding SINR is given as \cite{9119462}:
\begin{equation}
\mathrm{SINR}^T_{i,j}= \frac{P_{\mathrm{T}}{G^{(\mathrm{max})}_{\mathrm{tx}}\left(\theta\right)G^{(\mathrm{max})}_{\mathrm{rx}}\left(\theta\right)}\gamma_{T}}{N_{0}+I_{\mathrm{agg}}^{T}},
\label{eq: SINR_THZ}
\end{equation}
where $P_{T}$ is the transmit power of all the THz BSs. Furthermore, $\left(I_{\mathrm{agg}}^{T}\right)$ denotes the aggregate SINR at the $i$th user from the interfering THz BSs and is written as 

\begin{equation}
I_{\mathrm{agg}}^{T} \triangleq {\displaystyle \sum_{\begin{array}{c}
\forall k\in I,\forall m\in J\\
\{k,m\}\neq\left\{ i,j\right\} 
\end{array}}}P_{T} D_{k,m}h_{k,m},
\end{equation}
where $D_{i,j}$ represents the beam alignment between the $i$th user and $j$th BS and can take values as $\{G_{\mathrm{tx}}^{\left(\mathrm{max}\right)}G_{\mathrm{rx}}^{\left(\mathrm{max}\right)},G_{\mathrm{tx}}^{\left(\mathrm{max}\right)}G_{\mathrm{rx}}^{\left(\mathrm{min}\right)},G_{\mathrm{tx}}^{\left(\mathrm{min}\right)}G_{\mathrm{rx}}^{\left(\mathrm{max}\right)},G_{\mathrm{tx}}^{\left(\mathrm{min}\right)}G_{\mathrm{rx}}^{\left(\mathrm{min}\right)}\}$. The corresponding probability for each case is $F_{\mathrm{tx}}F_{\mathrm{rx}}$, $F_{\mathrm{tx}}(1-F_{\mathrm{rx}})$, $(1-F_{\mathrm{tx}})F_{\mathrm{rx}}$ and $(1-F_{\mathrm{tx}})(1-F_{\mathrm{rx}})$, where $F_{\mathrm{tx}} = \frac{|\theta_{\mathrm{tx}}|}{2\pi}$ and $F_{\mathrm{rx}} = \frac{|\theta_{\mathrm{rx}}|}{2\pi}$, respectively.

\subsection{Formulation as GAP}

When $i$th user is served by $j$th BS, its data rate is given as follows:
 \begin{equation}
 \mathrm{R}_{i,j}= W \log_{2}(1 + \mathrm{SINR}_{i,j}) \:,
\label{eq: Rij}
 \end{equation}
where ${W}$ denotes the available bandwidth, SINR between $i$th user and $j$th BS is represented by $\mathrm{SINR}_{i,j}$, which is determined using (\ref{eq: SINR_RF}) and (\ref{eq: SINR_THZ}) for RF and THz channels, respectively. Thus, the corresponding user-association problem which maximizes the sum rate ($R$) is formulated as the following GAP, given by

\begin{equation}
\begin{array}[b]{c}
{\displaystyle \text{maximize}}\quad R\triangleq\sum_{i=1}^{I}\sum_{j=1}^{J}u_{i,j}R_{i,j},\:\\
\!\!\!\!\!\!\!\!\!\!\!\!\!\!\!\!\!\!\!\!\!\!\!\!\!\!\!\!\!\textrm{subject to C1, C2 and C3,}
\end{array}
\label{eq: GAP_user}
\end{equation}
where $c_{j}=N^{b}$, $w_{i,j}=1$ $\forall i\in \mathcal{I},\, j\in \mathcal{J}$ in (\ref{eq: optimization_problem}) and $N^{b}$ represents a \textit{BS quota} i.e., the maximum number of users which can be served by each BS. Now, subsequently, we employ our proposed DUL based approach to solve (\ref{eq: GAP_user}) and analyze performance of the proposed approach using numerical examples.
\subsection{Parameter Settings}
Unless stated otherwise, the simulation parameters which are used to generate the feature vectors (or training and test data sets) are listed herein. Users are distributed within a circular disc of radius $100$ m. The molecular absorption coefficient $k_{a}$ is set as 0.05 m$^{-1}$ with 1\% of water vapor molecules. The absorption value is chosen from the realistic database and its corresponding central frequencies is $1.0$ THz \cite{jornet2011channel},  \cite{rothman2009hitran}. Without loss of generality, we normalize the sum  rate with the transmission bandwidth. The RF transmission frequency is set as $2.1$ GHz and  $\alpha$ = 2.5. The antenna gains $G_{\mathrm{tx}}^T$ and $G_{\mathrm{rx}}^T$ are taken as 25 dB. The antenna gains of RF transmitters and receivers are set as 0 dB. The transmit powers of all BSs are  taken as 1 W and  $N_{0}$ is -70~dBm. 

Unless specified otherwise, the number of epochs, the batch size, penalty parameter $\lambda$ and learning rate for our first scenario, 4 users and 4 BS, are taken as $50$, $128$, $6$ and $0.0001$, respectively. For our second scenario of 16 users and 4 BS the epochs and $\lambda$ are increased to $100$ and $10$, while the batch size and learning rate remain the same. For our first scenario we have an input and output vector of size $16$ and the number of neurons from the first hidden to the last hidden layer are given as $\left\{64,128,256,512,1024,2048\right\}$. For our second scenario we have an input and output vector of size $64$ and $\left\{128,256,512,1024,2048,2048,4096,4096\right\}$ neurons from the first hidden layer to the last. The remaining hyper-parameters are given in each figure. With these hyper-parameters, the results for a system with 4 BSs and 4 users are demonstrated in Fig. 2 to Fig. 4. Moreover, in all the examples, DNN is first trained for given hyper-parameters and then the trained DNN is used to obtain the sum data rate, averaged over $1000$ examples of the test data set.
Overall, $10000$ and $16000$ independent feature vectors were generated for the training data sets for 4 users and 16 users, respectively. In both cases, $1000$ were generated for testing.

 \begin{figure*}
\begin{minipage}{0.48\textwidth}
\centering
    \includegraphics[width=0.9\linewidth]{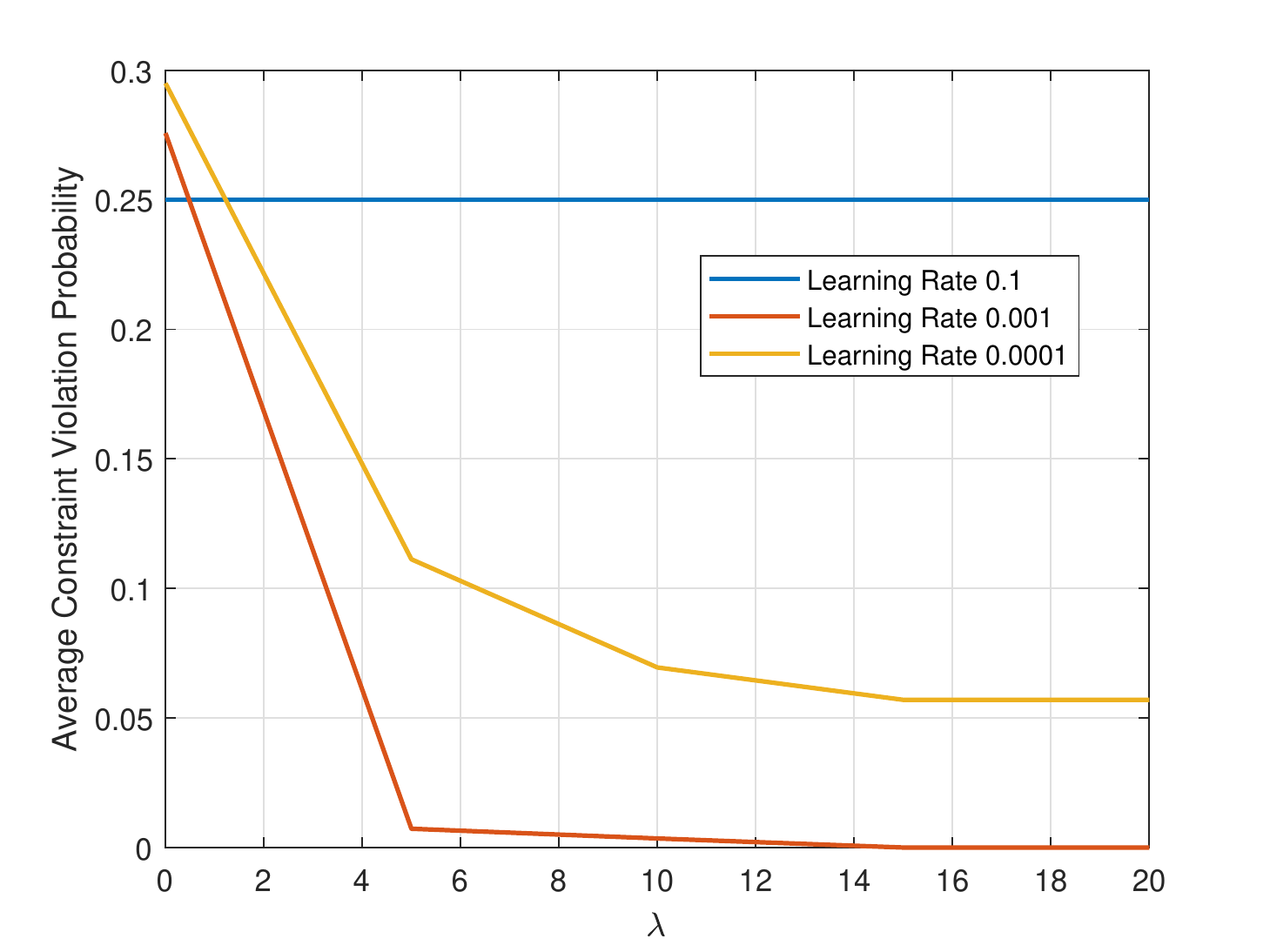}  
    \caption{Average constraint violation probability as a function of the penalty parameter $\lambda$  with different learning rates for 4 users and 4 BSs}
   \label{fig:lambda_vs_success}
\end{minipage}
\hfill
\begin{minipage}{0.48\textwidth}
   \centering
    \includegraphics[width=0.9\linewidth]{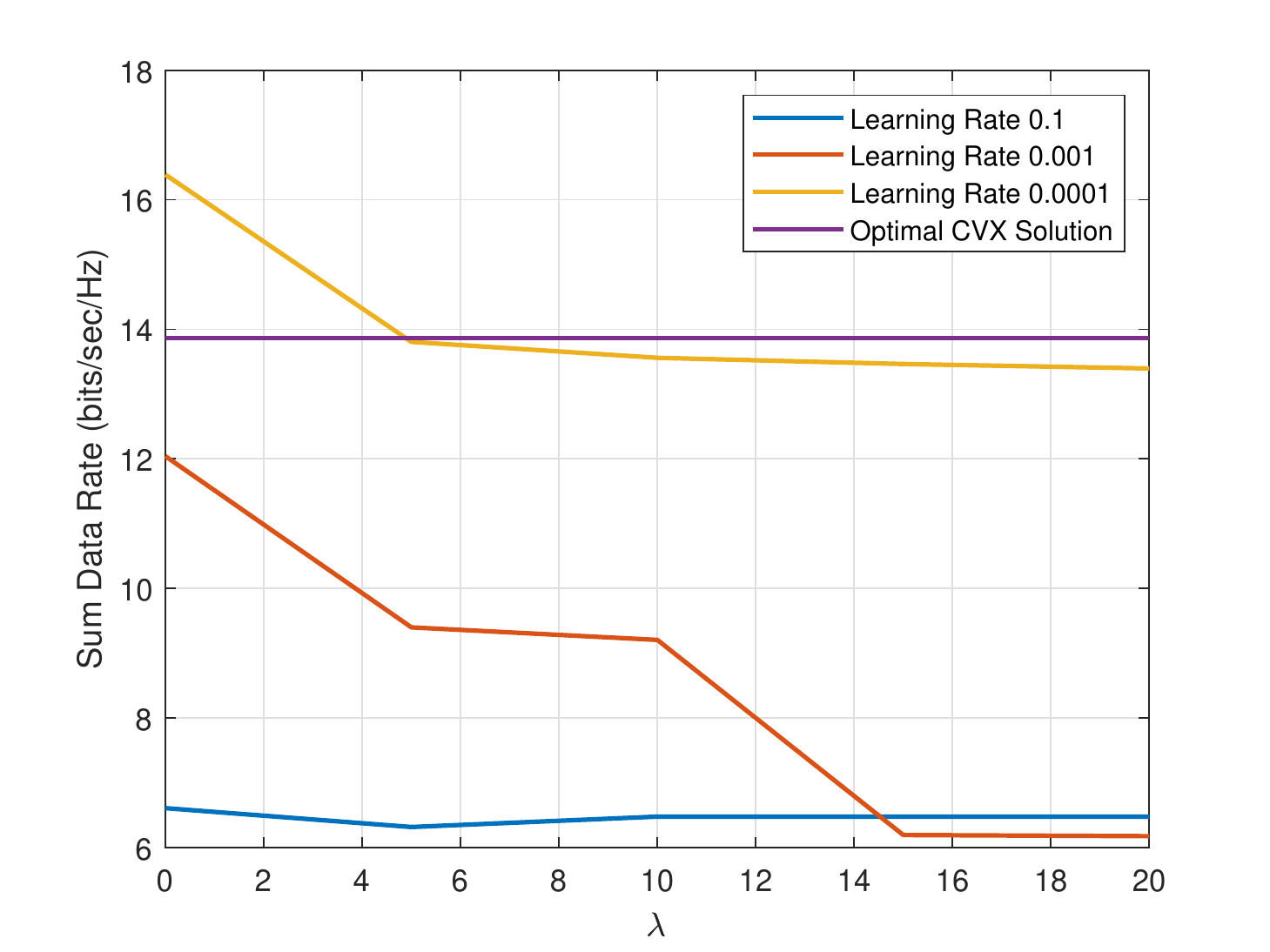}  
    \caption{Sum data rate as a function of $\lambda$ with different learning rates for 4 users and 4 BSs}
   \label{fig:lambda_vs_datarate}
   \end{minipage}
\end{figure*}

\begin{figure*}[h]
\begin{minipage}{0.45\linewidth}
        \includegraphics[width=0.95\linewidth]{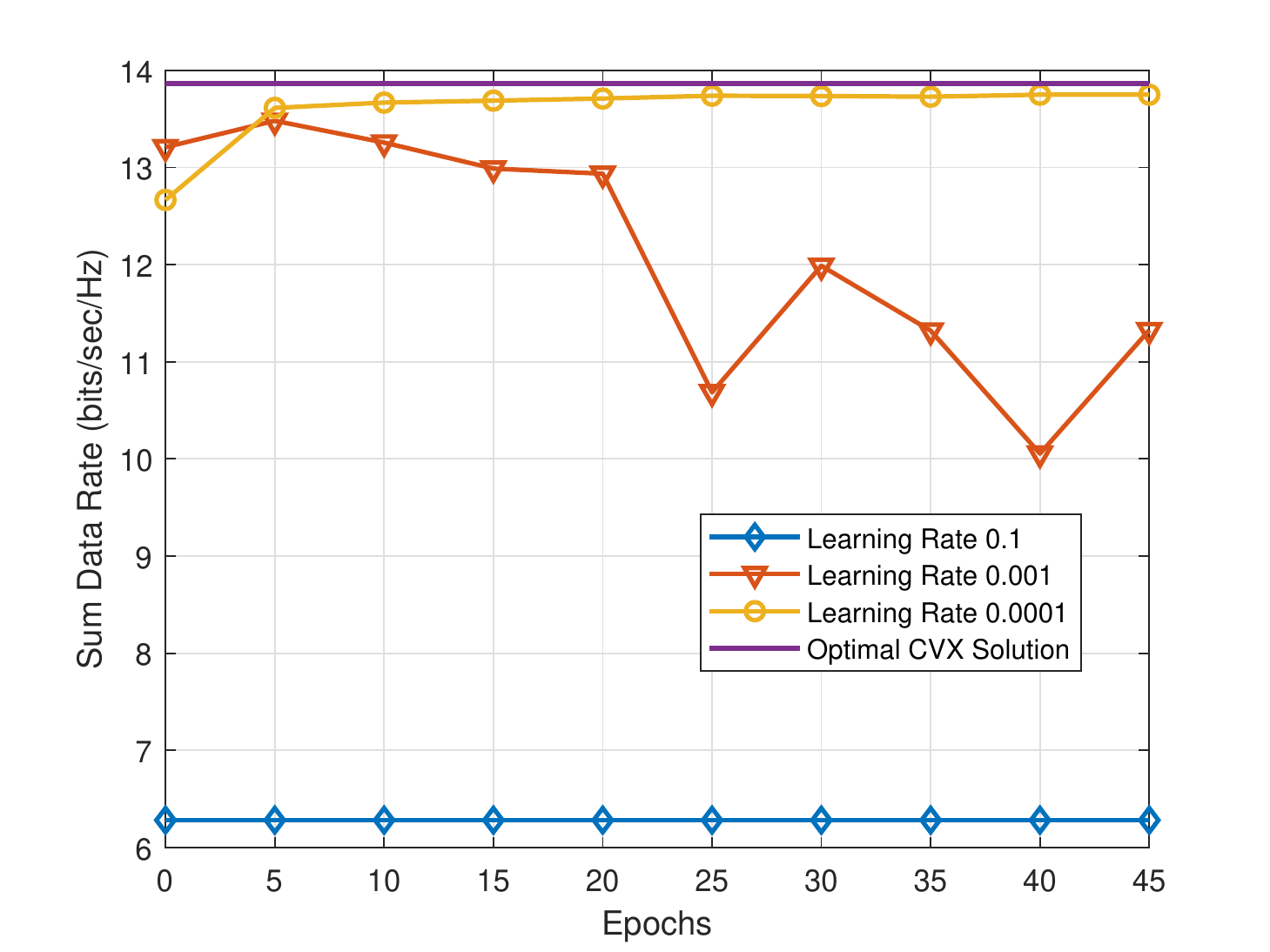}  
    \caption{Sum data rate as a function of the number of epochs  with batch size = 128, $\left(\lambda=6\right)$ for 4 users and 4 BSs}
   \label{fig:batch_vs_data_rate}
\end{minipage}
\hfill
   \hfill
   \begin{minipage}{0.45\linewidth}
    \includegraphics[width=0.95\linewidth]{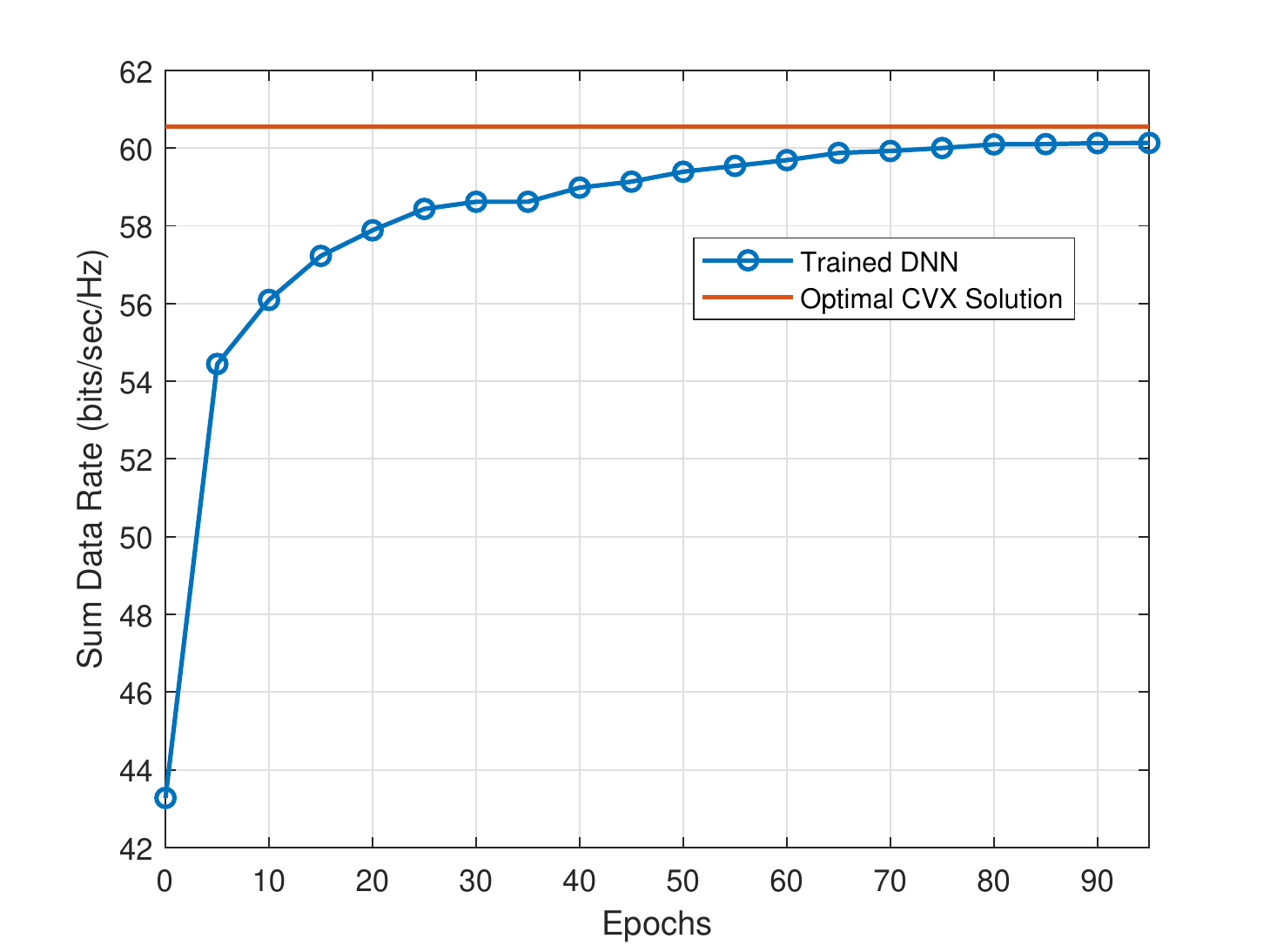}  
    \caption{Sum data rate as a function of number of epochs with batch size = 128, $\left(\lambda=10\right)$ for 16 users and 4 BSs}
   \label{fig:constraint_3_scalability}
   \end{minipage}
\end{figure*}


\begin{table*}
\centering
\caption{ Comparative analysis of the proposed DNN with the optimal solution.} 
{\begin{tabular}{|c|c|c|c|}
\hline
  \textbf{Schemes}&\textbf{ 4 Users} & \textbf{16 Users} & \textbf{Time Complexity (for 16 users)}\\ \hline\hline
  Optimal using CVX &  13.8625  &60.5487& $\sim$ 250 ms\\ \hline \hline
  DNN (with quota constraint) & 13.68 ($98.68\%$)&
   60.47 ($99.87\%$) & $\sim$ 0.24 ms\\ \hline\hline
\end{tabular}}
\label{Notation_Summary_mmwave}
\end{table*}

 \subsection{Optimal Solution - Benchmark}
 As the DNN provides a continuous user association profile i.e., $0 \leq u_{i,j} \leq 1, \:\forall i\in \mathcal{I},\,\forall j\in \mathcal{J}$, the performance of the proposed DUL approach is compared with the benchmark scheme in which the problem $\mathcal{P}$ is solved optimally using CVX by relaxing the binary constraint. Note that this relaxation can allow users to be associated to multiple BSs; however, this can be interpreted as {\em association probability} or a partial time allocation at each BS. For example, a fraction of 0.8 at one BS depicts that user associates to it 80\% of the time. 
 

\subsection{Proposed DUL Framework - Results}
In this subsection, we first present the simulation parameters and hyper-parameter settings of the proposed DNN, then we describe considered benchmark algorithms, and finally we present our main results and discussions. The DNN was trained and tested within Python using a TensorFlow backend. 


 A system with $4$ BSs and $4$ users is first considered. Then, we show the scalability of our proposed scheme with an example of 4 BSs and 16 users.
 
 In Fig. 2, first the DNN is trained for different values of penalty parameter $\lambda$  and learning rates, then the average constraint violation probability is plotted using the trained DNN.  Similarly, sum data rate is plotted for various values of $\lambda$ in Fig. 3. The trade-off between average constraint violation probability and sum data rate is controlled by the parameter $\lambda$ and thus choosing an optimal value has a strong impact on the overall performance. We note that both constraint violation probability and sum date rate decrease with increasing $\lambda$, as expected. From these figures, an appropriate value of $\lambda$ and learning rate is chosen. 
 
 Using $\lambda =6$, sum data rates are shown in Fig. 4 for different values of epochs and learning rates. By observing these figures (i.e., Fig. 2 to Fig. 4) collectively, it is evident that the suitable values of the hyper-parameters for 4 BSs and 4 users system are as follows:  learning rate=$0.0001$, number of epochs=$50$ and batch size=$128$. With these hyper-parameters of the trained DNN, the performance of the proposed DNN is compared with the optimal CVX scheme in Table I. Table I shows that the sum data rate of trained DNN (with $\lambda=6$) preforms very closely to the optimal CVX solution as the proposed DUL scheme. Our scheme achieves a sum data rate (i.e., $13.68$ bits/sec/Hz which is $98.68\%$ of the optimal value) with an average constraint violation probability of only $0.094$.





{ Table I depicts that the proposed DNN algorithm achieves close-to-optimal performance (i.e., sum data rates) as compared to the optimal CVX solution for constrained user-association problems.}  In terms of time complexity, it is evident that our unsupervised DNN approach outperforms the optimal CVX solution by a significant margin of $\sim$ 250 ms on average. This comparison clearly shows that with an increased number of users and BSs, the DNN's time complexity will be much lower compared to that of the optimal CVX solution. It is noteworthy that, in practice, where the channel coherence time is in the order of few milliseconds, the optimal CVX solution becomes impractical. On the other hand, our DUL framework  serves as a good solution under these circumstances.

Next, the number of users is increased to $16$ in Fig.~5. In order to train the DNN properly, the size of training set and number of epochs are increased to $16000$ feature vectors and $100$, respectively. Additionally, the amount of hidden layers are increased by $2$ as mentioned in section $C$ \textit{Parameter Settings}. With these settings, the sum data rate is given in Fig. 5. Table I displays the results for the 16 users scenario, we compare the performance of the trained DNN with the benchmark schemes. By taking $\lambda$ and learning rate equal to $10$ and $ 0.0001$, respectively, an average constraint violation probability of $0.094$ is achieved and sum data rate equal to $60.47$ bits/sec/Hz (which is $99.87\%$ of the  optimal value). The difference between the accuracy of 16 users and 4 users scenarios can be explained by considering the increase in network architecture complexity, the number of epochs, and the number of training samples.


\section{Conclusion}
In this paper, we proposed a new DUL approach for solving a classical GAP. In particular, we showed that a DNN can be trained to learn any GAP with the help of a customized loss function. With an aim to simplify a loss function and in order to make sure that the equality constraint is always satisfied, we proposed to use a $\mathsf{Softmax}$ function along with tensor splitting at the output of DNN. On the other hand, an inequality constraint is handled through a penalty parameter which is treated as one of the hyper-parameters of the DNN. This hyper-parameter is configured to achieve a trade-off between maximizing the objective function and meeting an inequality constraint.   Additionally, we formulated a user-association problem in the form of GAP and solved it using our DUL approach. Furthermore, for future directions of this research, extrapolating results for larger results will be done. Overall, numerical results demonstrate that the proposed approach yields close to optimal results and has approximately 1000 times lower time complexity, as compared to the optimal solution obtained by CVX.

\bibliographystyle{IEEEtran}
	\bibliography{main}

\end{document}